\documentclass[twoside,11pt]{article}

\usepackage{jmlr2e}

\jmlrheading{1}{2000}{1-48}{4/00}{10/00}{Marina Meil\u{a} and Michael I. Jordan}

\ShortHeadings{ParallelPC}{Thuc D. Le et al.}
\firstpageno{1}

\begin{document}

\title{ParallelPC: an R package for efficient constraint based causal exploration}

\author{\name Thuc D.\ Le \email Thuc.Le@unisa.edu.au \\
       \name Tao \ Hoang \email hoatn002@mymail.unisa.edu.au\\
       \name Jiuyong Li \email Jiuyong.Li@unisa.edu.au \\
       \name Lin Liu \email Lin.Liu@unisa.edu.au \\
       \name Shu \ Hu \email huysy011@mymail.unisa.edu.au \\
       \addr School of Information Technology and Mathematical Sciences\\
       University of South Australia\\
       Mawson Lakes, SA, Australia}

\editor{}

\maketitle

\begin{abstract}
Discovering causal relationships from data is the ultimate goal of many research areas. Constraint based causal exploration algorithms, such as  PC, FCI, RFCI, PC-simple, IDA and Joint-IDA have achieved significant progress and have many applications. A common problem with these methods is the high computational complexity, which hinders their applications in real world high dimensional datasets, e.g gene expression datasets. In this paper, we present an R package, \textit{ParallelPC}, that includes the parallelised versions of these causal exploration algorithms. The parallelised algorithms help speed up the procedure of experimenting big datasets and reduce the memory used when running the algorithms. The package is not only suitable for super-computers or clusters, but also convenient for researchers using personal computers with multi core CPUs. Our experiment results on real world  datasets show that using the parallelised algorithms it is now practical to explore causal relationships in high dimensional datasets with thousands of variables in a single multicore computer. \textit{ParallelPC} is available in CRAN repository at https://cran.r-project.org/web/packages/ParallelPC/index.html.
\end{abstract}


\section{Introduction}

Inferring causal relationships between variables is an ultimate goal of many research areas, e.g. investigating the causes of cancer, finding the factors affecting life expectancy. Therefore, it is important to develop tools for causal exploration from real world datasets.

One of the most advanced theories with widespread recognition in discovering causality is the Causal Bayesian Network (CBN),  \cite{pearl2009causality}. In this framework, causal relationships are represented with a Directed Acyclic Graph (DAG). There are two main approaches for learning the DAG from data: the search and score approach, and the constraint based approach. While the search and score approach raises an NP-hard problem, the complexity of the constraint based approach is exponential to the number of variables. Constraint based approach for causality discovery has been advanced in the last decade and has been shown to be useful in some real world applications. The approach includes causal structure learning methods, e.g. PC (\cite{Spirtes2000}), FCI and RFCI (\cite{colombo2012}), causal inference methods, e.g. IDA (\cite{maathuis2009estimating}) and Joint-IDA (\cite{nandy2014}), and local causal structure learning such as PC-Simple ( \cite{buhlmann2010}). However, the high computational complexity has hindered the applications of causal discovery approaches to high dimensional datasets, e.g. gene expression datasets where the number of genes (variables) is large and the number of samples is normally small.



In \cite{le2015fast}, we presented a method that is based on parallel computing technique to speed up the  PC algorithm. Here in the \textit{ParallelPC} package, we  parallelise a family of causal structure learning and causal inference methods, including PC, FCI, RFCI, PC-simple, IDA, and Joint-IDA. We also collate 12 different conditional independence (CI) tests that can be used in these algorithms. The algorithms in this package return the same results as those in the $pcalg$ package,  \cite{kalisch2012}, but the runtime is much lower depending on the number of cores CPU specified by users. Our experiment results show that with the \textit{ParallelPC} package it is now practical to apply those methods to high dimensional datasets in a modern personal computer.

\section{Contraint based  algorithms and their parallelised versions}

\begin{figure}[h]
\begin{centering}
\includegraphics[width = 4.5in, height=3.0in]{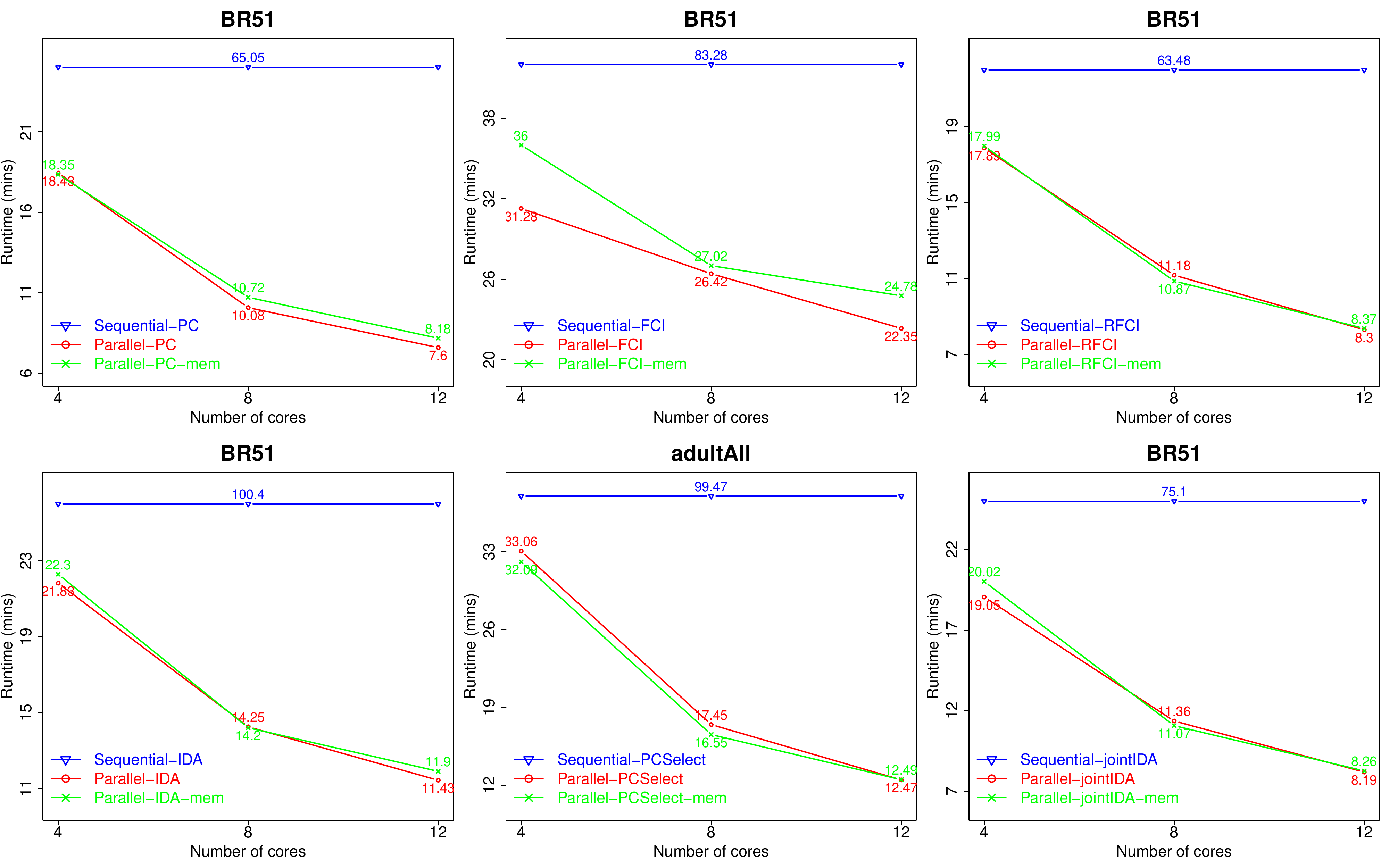}
\caption{Runtime of the sequential and parallelised versions (with and without the memory efficient option) of PC, FCI, RFCI, IDA, PC-simple, and Joint-IDA}
\label{PC}
\end{centering}
\end{figure}

We paralellised the following causal discovery and inference algorithms.

\begin{itemize}

\item PC, \cite{Spirtes2000}. The PC algorithm  is the state of the art method in constraint based approach for learning causal structures from data. It has two main steps. In the first step, it learns from data a skeleton  graph, which contains only  undirected edges. In the second step, it orients the undirected edges to form an equivalence class of DAGs.  In  the skeleton learning step, the PC algorithm starts with the fully connected network and uses the CI tests to decide if an edge is removed or retained.  The stable version of the PC algorithm (the Stable-PC algorithm, \cite{colombo2014}) updates the graph at the end of each level (the size of the conditioning set) of the algorithm rather than after each CI test. Stable-PC limits the problem of the PC algorithm, which is dependent on the order of the CI tests.  It is not possible to parallelise the Stable-PC algorithm globally, as the CI tests across different levels in the Stable-PC algorithm are dependent to one another. In \cite{le2015fast}, we proposed the Parallel-PC algorithm which parallelised the CI tests inside each level of the Stable-PC algorithm. The Parallel-PC algorithm is more efficient and returns the same results as that of the Stable-PC algorithm (see Figure 1).


\item FCI, \cite{colombo2012}. FCI is designed for learning the causal structure that takes latent variables into consideration. In real world datasets, there are often unmeasured variables and they will affect the leant causal structure. FCI was implemented in the $pcalg$ package and it uses PC algorithm as the first step. The skeleton of the causal structure learnt by PC algorithm will be refined by performing more CI test. Therefore, FCI is not efficient for large datasets.

\item RFCI, \cite{colombo2012}. RFCI is an improvement of the FCI algorithm to speed up the running time when the underlying graph is sparse. However, our experiment results show that it is still impractical for high dimensional datasets.

\item PC-simple, \cite{buhlmann2010}. PC-simple is a local causal discovery algorithm to search for parents and children of the target variable. In dense datasets where the target variable has large number of causes and effects, the algorithm is not efficient. We utilise the idea of taking order-independent approach (\cite{le2015fast}) on the local structure learning problem to parallelise the PC-simple algorithm. 

\item IDA , \cite{maathuis2009estimating}. IDA is a causal inference method which infers the causal effect that a variable has on another variable. It firstly learns the causal structure from data, and then based on the  learnt causal structure, it estimates the causal effect between a cause node and an effect node  by adjusting the effects of the parents of the cause.  Learning the causal structure is time consuming. Therefore, IDA will be efficient when the causal structure learning step is improved. Figure 1 shows that our parallelised version of the IDA improves the efficiency of the IDA algorithm significantly.

\item Joint-IDA, \cite{nandy2014}. Joint-IDA estimates the effect of the target variable when jointly intervening a group of variables. Similar to IDA, Joint-IDA learns the causal structure from data and the effects of intervening multiple variables are estimated.
\end{itemize}

To illustrate the effectiveness of the parallelised algorithms, we apply the sequential and parallelised versions of PC, FCI, RFCI, IDA and Joint-IDA algorithms to a breast cancer gene expression dataset. The dataset includes  50 expression samples with 92 microRNAs (a class of gene regulators) and 1500 messenger RNAs that was used to infer the relationships between miRNAs and mRNAs downloaded from \cite{le2015mirna}. As PC-simple (PC-Select) is efficient in small datasets, we use the Adult dataset from UCI Machine Learning Repository with 48842 samples. We use the binary discretised version from \cite{li2015} and select 100 binary variables for the experiment with PC-simple. We run all the experiments on a Linux server with 2.7 GB memory and 2.6 Ghz per core CPU.

As shown in Figure 1, the parallelised versions of the algorithms are much more efficient than the sequential versions as expected, while they are still generating the same results.

The parallelised algorithms also detect the free memory of the running computer to estimate the number of CI tests that will be distributed evenly to the cores. This step is to ensure that each core of the computer will not hold off a big amount of memory while waiting for the synchronisation step. The memory-effcient procedure may consume a little bit more time compared to the original parallel version. However, this option is recommended for computers with limited memory resources or for big datasets.



\section{Conditional independence tests for constraint based methods}

It is shown that different CI tests may lead to different results for a particular constraint based algorithm, and a CI test may be suitable for a certain type of datasets. In this package, we collate 12 CI tests in the $pcalg$ (\cite{kalisch2012}) and $bnlearn$ (\cite{scutari2009}) packages to provide options for function calls of the constraint-based methods. These CI tests can be used separately for the purpose of testing (conditional) dependency between variables. They can also be used within the constraint based algorithms (both sequential and parallelised algorithms) in the \textit{ParallelPC} package. The following codes show an example of running the FCI algorithm using the sequential version in the $pcalg$ package, the parallelised versions with or without the memory efficient option, and using a different CI test (mutual information) rather than the Gaussian CI test.
\begin{verbatim}
  ## Using the FCI-stable algorithm in the pcalg package
     library(pcalg)
     data("gmG")
     p<-ncol(gmG$x)
     suffStat<-list(C=cor(gmG$x),n=nrow(gmG$x))
     fci_stable(suffStat, indepTest=gaussCItest, p=p, skel.method="stable", alpha=0.01)
  ## Using fci_parallel without the memory efficient option
     fci_parallel(suffStat, indepTest=gaussCItest, p=p, skel.method="parallel",
        alpha=0.01, num.cores=2)
  ## Using fci_parallel with the memory efficient option
     fci_parallel(suffStat, indepTest=gaussCItest, p=p, skel.method="parallel",
        alpha=0.01, num.cores=2, mem.efficient=TRUE)
  ## Using fci_parallel with mutual information test
     fci_parallel(gmG$x, indepTest=mig, p=p, skel.method="parallel",
        alpha=0.01, num.cores=2, mem.efficient=TRUE)
\end{verbatim}
\acks{This work has been supported by Australian Research
Council Discovery Project DP140103617. }






\vskip 0.2in
\bibliography{sample}

\end{document}